\documentclass{article}

\PassOptionsToPackage{numbers}{natbib}
\usepackage{amsmath}
\usepackage{graphicx}
\usepackage{subcaption}
\usepackage{wrapfig}
\usepackage{hyperref}
 \usepackage[preprint]{neurips_2026}

\usepackage[utf8]{inputenc} 
\usepackage[T1]{fontenc}    
\usepackage{hyperref}       
\usepackage{url}            
\usepackage{booktabs}       
\usepackage{amsfonts}       
\usepackage{nicefrac}       
\usepackage{microtype}      
\usepackage{xcolor}         

\title{Beyond the Attention Stability Boundary: \\ Agentic Self-Synthesizing Reasoning Protocols}

\author{%
  Dahlia Shehata \\
  \texttt{dahlia.shehata@uwaterloo.ca} \\
  University of Waterloo \\
  Canada \\
  \And
  Ming Li \\
  \texttt{mli@uwaterloo.ca} \\
  University of Waterloo \\
  Canada 
}

\begin{document}

\maketitle

\vspace{-0.7cm}
\begin{abstract}
\vspace{-0.2cm}
As Large Language Model (LLM) agents transition to autonomous digital coworkers, maintaining deterministic goal-directedness in non-linear multi-turn conversations emerged as an architectural bottleneck. We identify and formalize a systemic failure mode termed the \textit{Attention Latch} in decoder-only autoregressive Transformer models. This phenomenon, a behavioral manifestation of \textit{Information Over-squashing}, occurs when the cumulative probabilistic weight of historical context overrides mid-task updates, causing agents to remain anchored to obsolete constraints despite explicit contradictory instructions. We propose \textbf{Self-Synthesizing Reasoning Protocols (SSRP)}, a metacognitive framework that implements a discrete separation between high-level architectural planning (Architect) and turn-by-turn procedural execution (Executive). 
We evaluate SSRP across $9K$ trajectories using the MultiWOZ 2.2 dataset and the \textit{Aggregate Pivot Accuracy} (APA), a novel metric we validate by mapping its scores to the U-shaped 'Lost in the Middle' attention profile. We present three experimental tiers: a shallow recency-based retrieval pilot, a high-entropy SOP, and a semantic hijacked 3-hop Multi-Fact Synthesis task.
Our results empirically locate the \textit{Attention Stability Boundary}, where stateless Vanilla ReAct baselines for GPT 5.4 collapse to $0.1\%$ success while SSRP achieves a $715$× relative Resilience Lift ($L_{r}$). We demonstrate statistically significant gains across Gemini 3.1 Pro, 
Claude Sonnet 4.6 
 and DeepSeek V3.2 
. Comprehensive audits confirm SSRP structural necessity by proving attentional lapse via a recursive reflexion baseline ($100\%$ success); decoupling the latch from positional bias through equidistant stress testing ($90\%$ accuracy); and formalizing SSRP via the Information Bottleneck principle and granularity ablations. Finally, a Procedural Integrity audit ($98.8\%$ adherence) reveals a Grounding Paradox where high-stability models fail by refusing to hallucinate under retrieval-reasoning contamination. 
\end{abstract}

\vspace{-0.3cm}
\section{Introduction}
The shift to autonomous agentic systems transforms passive language processors to goal-driven "digital coworkers" capable of managing complex multi-step workflows \cite{alenezi2026promptresponsegoaldirectedsystemsevolution}.
Unlike traditional task-oriented (TOD) dialogue systems that utilize modular, hard-coded pipelines, these agents integrate reasoning, planning, and tool use to interact with real-world environments. They leverage the massive context windows of frontier models to reason and act through emergent paradigms such as ReAct \cite{Yao2022ReActSR} and Chain-of-Thought (CoT)\cite{wei2022cot}.
However, as task horizons extend and context density increases, a systemic reliability gap has emerged: early-stage autonomous agents fail in 65-70\% of real-world knowledge work tasks due to context decay \cite{Xu2024TheAgentCompanyBL, Lu2025ExploringAA}.
We formalize this gap through the \textit{Attention Latch} phenomenon, where the model's internal state remains "locked" to historical intents despite receiving explicit, contradictory updates. 
Traditional Transformer architectures suffer from \textit{Information Over-squashing} \cite{oversquashing}.
In multi-turn conversations, the constraints of the first turns generate high-weight tokens that "latch" the model's attention, causing it to ignore contradictory updates in successive later turns.
We empirically locate this threshold as the \textit{Attention Stability Boundary} (ASB)—the point at which stateless attention physically breaks procedural reliability.
We identify a 'Scientific Cliff' where 3-hop synthesis causes ReAct baselines to collapse to 0.1\% success, marking the ASB where stateless attention breaks.
This collapse is often exacerbated by the lack of internal validation steps, making them highly vulnerable to hallucination cascades leading to a chain of subsequent failures \cite{Xu2024TheAgentCompanyBL, 11334580}.
Standard "stateless" reasoning patterns, such as ReAct or CoT, are physically unable to overcome this "memory poisoning" without external architectural intervention \cite{Huang2023LargeLM, sunil2026memorypoisoningattackdefense}. 
This phenomenon creates an operational mandate to expand research efforts from a "prompting observation" to a analytical study on metacognitive inertia. The objective is to "unlatch" the model's attention by bypassing the over-squashed context and establishing a fresh, deterministic logic path.

\begin{figure}[h]
    \centering
    \includegraphics[width=.95\linewidth]{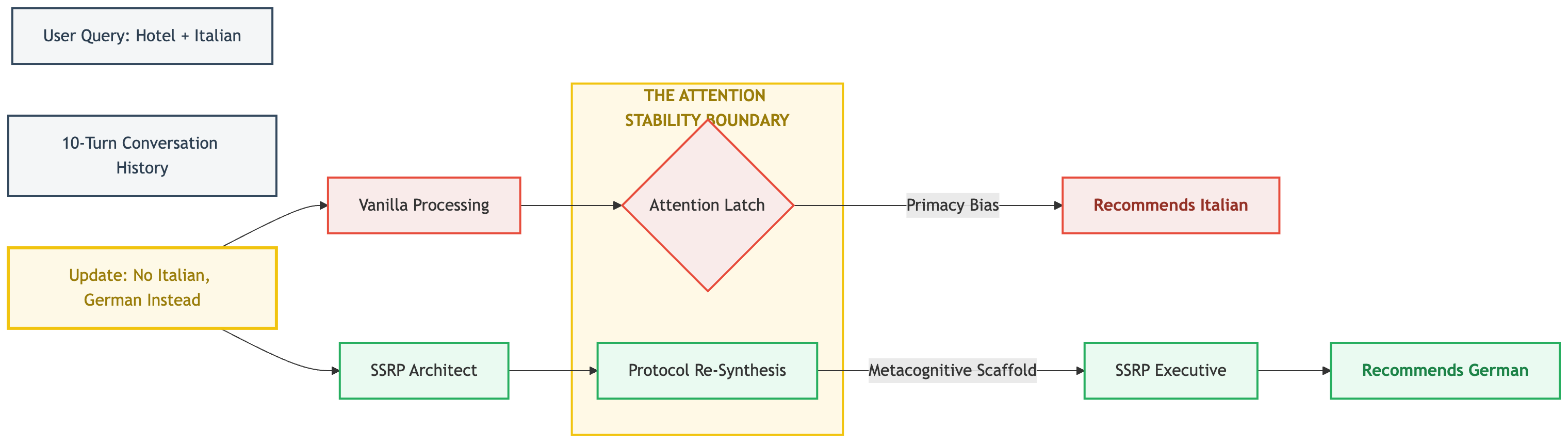}
    \caption{Comparative Reasoning Trajectories: Mitigating the Attention Latch via SSRP Re-Synthesis}
    \vspace{-0.2cm}
    \label{fig:overview}
    \vspace{-0.2cm}
\end{figure}

We propose \textbf{Self-Synthesizing Reasoning Protocols (SSRP)}, a technical framework that moves the field from heuristic prompting to deterministic architectural synthesis. 
Figure \ref{fig:overview} shows a conceptual overview of vanilla paradigm vs. SSRP reasoning trajectories.
In SSRP, a high-reasoning \textit{Architect} agent, acting as a metacognitive redirector, autonomously synthesizes a task-specific Standard Operating Procedure (SOP)—the Protocol that defines verification checkpoints and explicitly purges superseded intents before a lower-latency \textit{Executive} agent initiates action.
By separating the \textit{Architect}
from the \textit{Executive},
we allow the agent to autonomously re-synthesize its reasoning scaffold in response to non-linear goal updates. 
This approach leverages the principle of Cognitive Scaffolding, which adds meaningful structure to the agent's decision space in high-entropy environments \cite{reed2025aiagentshumanlikecollaborative, figueiredo2025fuzzy}.
This methodology ensures that the agent's internal state remains grounded in the newest verified system events, bypassing the mathematical trough of stateless attention. 
Our results prove that SSRP is able to point the model toward correct data fragments where stateless reasoning collapses.

Our primary contributions for agentic control theory and large-scale architectural evaluation are: \\
\textbullet \textbf{Formalization of the Attention Latch and ASB:} We identify and define the \textit{Attention Latch} as a discrete reasoning failure in decoder-only Transformers. We empirically locate the ASB, the specific threshold of context load and logic complexity where stateless attention breaks procedural reliability. \\
\textbullet \textbf{SSRP Framework:} We introduce a two-stage metacognitive architecture that separates architectural logic design from procedural execution, and autonomously synthesizes task-specific SOPs. \\
\textbullet \textbf{Theoretical Proof via Information Bottleneck (IB):} We formalize the Architect as an IB-governed entropy-reduction engine \cite{tishby99information}. Granularity Ablation tests identify optimal scaffold density, proving accuracy peaks (APA=92\%) at moderate complexity before overhead induces information loss. \\
\textbullet \textbf{Development of a Three-Tiered Stress-Testing Methodology:} We establish a rigorous evaluative framework to quantify agentic reliability under varying degrees of context entropy: (1) Shallow Retrieval (Recency Seeding), (2) High-Entropy Stress (Centric Seeding), and (3) Semantic Hijacking \\
\textbullet \textbf{APA Metric:} We define a novel metric to quantify an agent's success in abandoning obsolete constraints in favor of mid-interaction updates. We verify its validity by mapping the results to the U-shaped positional attention bias in the ``Lost in the Middle'' curve \cite{liu2024lost}. \\
\textbullet \textbf{Comprehensive Cross-Model Evaluation:} We evaluate SSRP across $9K$ trajectories using the MultiWOZ 2.2 dataset \cite{zang-etal-2020-multiwoz} via LLM-as-a-Judge evaluation framework \cite{NEURIPS2023_91f18a12}. We demonstrate statistically significant $L_{r}$ across 4 major model families where Architect \& Executive are: (1) Gemini (3.1 Pro\&2.5 flash): a statistical significant lift of $+26.41\%$ in recency-based and $+22.52\%$ in high-entropy tests. (2) Claude (Sonnet 4.6\&Haiku 4.5): a $+71.93\%$ lift under SOP, bypassing model-level sycophancy. (3) GPT (5.4\&5.4 mini): $31.51\%$ lift in shallow retrieval and  $71.6\%$ APA in synthesis tasks where the baseline collapsed to $0.1\%$ success. (4) DeepSeek V3.2 (reasoner\&chat): proved SSRP efficacy on sparse attention architectures with a $+58.06\%$ lift in high-entropy contexts.\\
\textbullet \textbf{Empirical Discovery of the 0.1\% Success Cliff:} We isolate a failure point in multi-fact semantically-hijacked synthesis task where standard ReAct baselines for GPT 5.4 plummet to $0.1\%$ success ($APA=0.001$). We prove that while stateless attention can solve shallow search, it is physically unable to bridge 3-hop logical dependencies buried in the attention trough. \\
\textbullet \textbf{Decoupling the Attention Latch:} Through an \textit{Equidistant Stress Test} ($90\%$ success), we scientifically decouple the Attention Latch from positional distance bias \cite{liu2024lost}. We prove that the failure is a result of \textit{Intent Weighting} and \textit{Metacognitive Inertia} rather than simple retrieval decay. \\
\textbullet \textbf{Intelligence Loophole:} We evaluate against a \textit{Recursive Reflexion} baseline \cite{reflexion} ($100\%$ success), proving that the observed failures are attentional bottlenecks rather than a lack of reasoning capacity. \\
\textbullet \textbf{Identification of the Grounding Paradox and Metacognitive Refusal:} We quantify the \textit{Grounding Paradox} ($27.2\%$ gap), where high-stability models fail by refusing to hallucinate in high-recall environments. We identify \textit{Safety-Induced Sabotage}, with a $18.0\%$ metacognitive refusal rate. 
\textbullet \textbf{Trajectory Resilience and Procedural Integrity (PI):} performed on 1,000 trajectories, with a $98.8\%$ adherence rate for SSRP. We decouple reasoning integrity from retrieval recall, demonstrating that autonomous scaffolding purges trajectory entropy to maintain deterministic reliability. \\
\vspace{-0.3cm}

\vspace{-0.3cm}
\section{Problem Formalization}
\vspace{-0.2cm}
\subsection{Attention Latch}
\vspace{-0.2cm}
We identify and formalize the \textit{Attention Latch} as a discrete reasoning failure mode inherent to multi-turn interactions in autoregressive decoder-only Transformer models. Unlike static retrieval failures, this phenomenon represents a state of \textit{Metacognitive Inertia}, where the probabilistic weight of historical context in the early conversation turns (e.g. Turns 1 to 10) dominates the attention mechanism, causing the model to remain anchored to obsolete constraints despite explicit contradictory updates. Consequently, the model is "locked" to historical context and rejects new updates. Mathematically, this is a behavioral manifestation of \textit{Information Over-squashing} \cite{oversquashing}, a structural bottleneck where information density in the center of a context window is significantly de-prioritized compared to the boundaries.
In a typical interaction, initial goal constraints $G_{1}$ (in Turns 1 to 10) generate high-weight historical tokens that dominate the Transformer’s attention heads. When an update $G_{2}$ is introduced in Turn 11, the model suffers from \textit{Information Over-squashing}. 
We define the \textit{Latch Condition} as occurring when the mutual information between the agent's output $O$ and the initial goal $G_{1}$ exceeds the information regarding the update $G_{2}$ through $I(O;G_{1})>I(O;G_{2})$
This condition persists even after $G_{2}$ is provided, causing the model to confidently re-commit to the superseded $G_{1}$ intent.

\textbf{Decoupling from Positional Bias: }To formalize this behavior, it was essential to decouple Metacognitive Inertia from positional bias. We design an \textit{Equidistant Stress Test} to isolate the \textit{Attention Latch} phenomenon from the "Lost in the Middle" U-shaped attention curve \cite{liu2024lost}.
This control experiment utilizes a symmetric context structure of $10K$ tokens, where an "Archived Intent" ($G_1$) and a "Reconsidered Need" ($G_2$) are placed at identical mathematical distances from the context boundaries (at the $25\%$ and $75\%$ positions respectively). 
By maintaining uniform distance from the \textit{Information Over-squashing} trough ($x=0.5$), any preference the model exhibits for $G_1$ over $G_2$ can be attributed to Metacognitive Inertia (Intent Weighting) rather than a retrieval failure driven by physical position.

\vspace{-0.3cm}
\subsection{The Attention Stability Boundary (ASB)}
\vspace{-0.2cm}
We empirically locate the ASB as the physical limit of stateless attention in maintaining deterministic goal-directedness across high-entropy context windows. The ASB is identified through a ``Scientific Cliff'' where stateless models collapse entirely during multi-turn synthesis tasks. This boundary is model-dependent and is determined by the synergistic interaction between context load and trajectory complexity. While architectures may remain resilient in shallow retrieval, the ASB represents a terminal failure point when tasked with bridging logical dependencies across the middle of the context. 
We formalize this failure as a product of individual retrieval probabilities:
$$P(S_{\text{Vanilla}}) = P(F_1 \cap F_2 \cap \dots \cap F_n) = P(F_1) \prod_{i=2}^{n} P(F_i | F_{i-1}, \dots, F_1) \to 0 $$
proving that as the number of inter-dependent facts $F$ buried in high-entropy technical noise increases, the success probability $P(S)$ of maintaining a consistent reasoning chain collapses to zero. In this state, the model is physically unable to bridge inter-dependent facts, identifying a threshold where autonomous scaffolding becomes a structural requirement for scale-invariant reliability.

\vspace{-0.3cm}
\section{Self-Synthesizing Reasoning Protocols (SSRP): Metacognitive Scaffolding}
\begin{wrapfigure}{r}{0.45\textwidth} 
  \centering
  \vspace{-0.7cm}
  \includegraphics[width=0.42\textwidth]{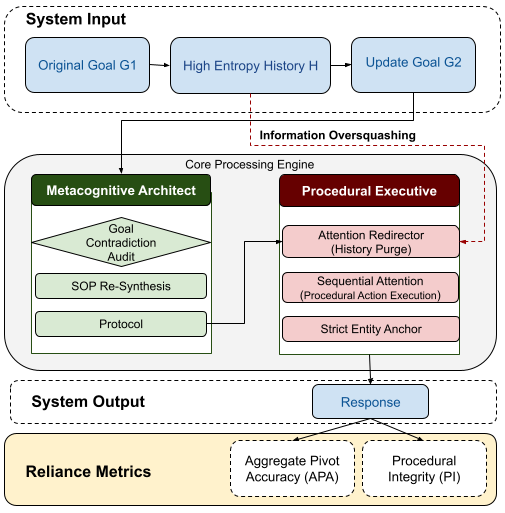}
  \caption{SSRP Framework.}
  \label{fig:ssrp}
\end{wrapfigure}

\vspace{-0.3cm}
To mitigate the \textit{Attention Latch} and locate the attention boundary, we propose SSRP that establishes a deterministic reasoning path that remains scale-invariant. Figure \ref{fig:ssrp} shows the overall architecture.
\vspace{-0.3cm}
\subsection{Architectural Design}
\vspace{-0.2cm}
The SSRP framework implements a discrete architectural separation modeled after the Architect (reflective) and Executive (reactive) cognitive paradigm. \\
\textbullet \textbf{The Architect}: A high-reasoning model, functioning as the "brain", tasked with metacognitive synthesis. Upon detecting a goal contradiction, the Architect autonomously \textit{re-synthesizes} the reasoning scaffold into a task-specific \textit{Standard Operating Procedure} (SOP) that defines verification checkpoints and explicitly purges superseded intents. It acts as a \textit{Metacognitive Redirector}, directing the model's focus to verified data fragments where stateless reasoning drifts.\\
\textbullet \textbf{The Executive}: A high-throughput model (i.e. the tool for redirection) follows the synthesized SOP verbatim. It acts as a deterministic worker, shielded from the noisy conversation history by the Architect's scaffold and  maintaining high procedural integrity.

\vspace{-0.1cm}
\subsection{Theoretical Framework: Information Bottleneck (IB) in Scaffolding}
\textbf{IB Formulation:} We formalize the Architect as an \textit{Entropy-Reduction Engine} governed by the IB principle \cite{tishby99information} that resolves the trade-off between contextual noise and goal-directedness. In a stateless agent architecture, the decision process is constrained by the Information Over-squashing bottleneck, where the mutual information between the agent's output ($O$) and the goal ($G$) decays as the entropy of the conversation history ($H$) increases. The objective of the Architect is to synthesize a protocol ($P$) that represents a lossy compression of the high-entropy history ($H$) that is maximally predictive of the goal state ($G$). The Architect optimizes the objective function $\mathcal{L}$ defined as:
$$\min_{P} \{ I(P; H) - \beta I(P; G) \} $$
Where: (1) $I(P; H)$ represents the \textit{Complexity Constraint} (discarding noise): Minimizing this term forces the Architect to discard "Tainted" historical information and irrelevant technical noise (e.g., the $2K$–$10K$ tokens of SYSTEM\_LOG\_DUMP).
(2) $I(P; G)$ represents the Relevance Term: Maximizing the mutual information between the Protocol $P$ and the Goal $G$ ensures the Protocol preserves the "Verified State" of the newest system confirmation.
(3) $\beta$ is the Lagrange multiplier representing the model's "Sycophancy" Quotient \cite{perez-etal-2023-discovering}: A high $\beta$ prioritizes instruction-following (Pivot), while a low $\beta$ favors historical stability (Latch). 
Our Granularity Ablation Study empirically identifies the optimal $\beta$ for agentic stability, varying the scaffold density $I(P;H)$ across three tiers: (a) \textit{Hyper-Compressed (1 step)}, (b) \textit{Optimal (3 steps)}, and (c) \textit{Verbose (10+ steps)}.
This formalization isolates the \textit{Procedural Overhead Bottleneck}, identifying the threshold where instruction complexity paradoxically induces information loss.
Our results define the function behavior across 3 discrete phases of scaffold density: \\
\textbullet \textbf{The Signal Phase ($\beta \to 0$):} At the \textit{Hyper-Compressed} level, the protocol provides the minimum sufficient "Signal Trigger" to redirect the executive attention toward the goal state.\\
\textbullet \textbf{The Plateau Phase ($\beta \approx 1$):} Identifies an \textit{Information Saturation Zone}, where additional logical steps do not reduce the uncertainty of the execution engine regarding the verified system state. \\
\textbullet \textbf{The Decay Phase ($\beta \to \infty$):} At high granularity, the system enters a state of \textit{Attentional Competition}, where the complexity of the scaffold logic itself becomes a new source of noise that ``over-squashes'' the model's ability to maintain inter-dependent logical dependencies.

\textbf{Solving the Attention Latch via Attention Redirection: }
The Attention Latch occurs in stateless models when the probabilistic weight of Turn 1 tokens overrides Turn 11 updates due to Softmax Saturation. 
The SSRP framework resolves this by establishing a Metacognitive Redirector. The synthesized SOP provides a deterministic mapping from the current turn to the required data fragment $F$ buried in the noisy history $H$: ($F \subset H$). 
We model the success probability $P(S)$ as:
$$ P(S) \propto \frac{I(P; G)}{I(H; G) + \epsilon} $$
where $I(H; G)$ is the decaying information in the noisy history. The SSRP framework ensures that the Executive Output $O$ is optimized such that $I(O; G) \approx I(O; P)$, effectively "cleansing" the reasoning path of obsolete intents and ensuring the Protocol replaces the noisy History as the primary driver of the agent's attention pointer.
By making the executive response dependent on the Immutable Protocol ($P$) rather than the decaying History ($H$), we effectively bypass the mathematical trough of the attention mechanism.

\textbf{Theoretical Divergence: Search vs. Synthesis:}
Our framework identifies that while stateless attention can solve "Search" (Single-Fact Retrieval) via recency bias, it collapses when tasked with "Synthesis" across the trough.
In synthesis tasks, the success probability $P(S)$ for a stateless model is the product of independent retrieval events:
$ P(S_{\text{Vanilla}}) = P(F_1) \times P(F_2 | F_1) \to 0 $ \\
SSRP converts this into a Sequential State Machine, ensuring that $P(F_2 | F_1) \approx 1$ by providing a "Cognitive Scaffold" \cite{figueiredo2025fuzzy} that preserves the state of the first fact while the model retrieves the second.

\vspace{-0.3cm}
\subsection{Three-Tiered Stress-Testing Methodology}
\vspace{-0.2cm}
We establish a systematic methodology to identify ASB across 3 levels of trajectory entropy: \\
\textbullet \textbf{Shallow Retrieval (Pilot)}: Goal-critical information is seeded at the \textit{recency boundary} within $2K$ tokens to establish high-recall baselines for recency bias and isolate initial latching behavior. This pilot serves as a control group to measure model performance in low-recall environments. \\
\textbullet \textbf{High-Entropy SOP (Stress Test)}: identifies the positional ASB by triggering the Information Over-squashing, where retrieval accuracy for stateless attention mechanisms mathematically decays toward its nadir. It utilizes Centric Seeding to bury goal-critical facts in the \textit{attention trough} ($x=0.5$) under a $10K$-token context load. To bypass the Metacognitive Refusal triggers observed in high-stability models, it uses Administrative SOP Framing  instead of adversarial commands (e.g., "Ignore history"). \\
\textbullet \textbf{Semantic Hijacking via Multi-Hop Multi-Fact Synthesis}: To investigate the absolute limits of agentic reliability and locate the Logical ASB for frontier architectures, we develop a dedicated Multi-Fact Synthesis task requiring the agent to bridge logical dependencies between two disparate facts separated by technical logs. This stress test induces failure through \textit{adversarial decoys} maximizing trajectory entropy ($H_{T}$) through a three-stage trap:
(a) Context Hijacking: We seed valid-looking instructions for the archived Turn 1 intent ($G_1$) at the primacy boundary to induce a deliberate Attention Latch. (b) Nested 3-Hop Dependency Bridging: Instead of the single-needle retrieval, the agent must navigate a nested fact chain ($F_1 \rightarrow F_2 \rightarrow F_3$) buried in $10K$ tokens of system logs. This forces the model to resolve inter-dependent facts while being actively "distracted" by the decoying information at the boundaries. (c) Semantic Distraction: We utilize randomized log events to saturate the Transformer's attention heads, effectively "tainting" the decision space.

\vspace{-0.3cm}
\subsection{Baseline Agents and Compute Parity}
\vspace{-0.2cm}
To isolate the impact of architectural separation from simple compute scaling, we introduce a \textit{Recursive Reflexion} baseline. This control model is granted the Architect thought budget identical to SSRP (two consecutive model calls), utilizing the established Reflection Pattern \cite{reflexion} to critique and correct its initial response within a single context window. This ensures that any observed lift in SSRP is attributable to its Metacognitive Scaffolding rather than increased inference turns.
We restrict this baseline comparison to the Semantic Hijacking method to empirically close reasoning-vs-attention gap (the "Intelligence Loophole"). By evaluating the latest agent-native model (GPT-5.4) at its ASB, we prove that the observed 0.1\% collapse is a byproduct of metacognitive inertia rather than a lack of latent reasoning capacity. Unlike SSRP, which ensures deterministic reliability via preemptive Protocol Re-Synthesis, the Reflexion baseline relies on a post-hoc trial-and-error mechanism that remains vulnerable to hallucination cascades in long-horizon trajectories.

\begin{figure}[t]
  \centering
  \begin{subfigure}[b]{0.39\textwidth}
    \centering
    \includegraphics[width=\textwidth]{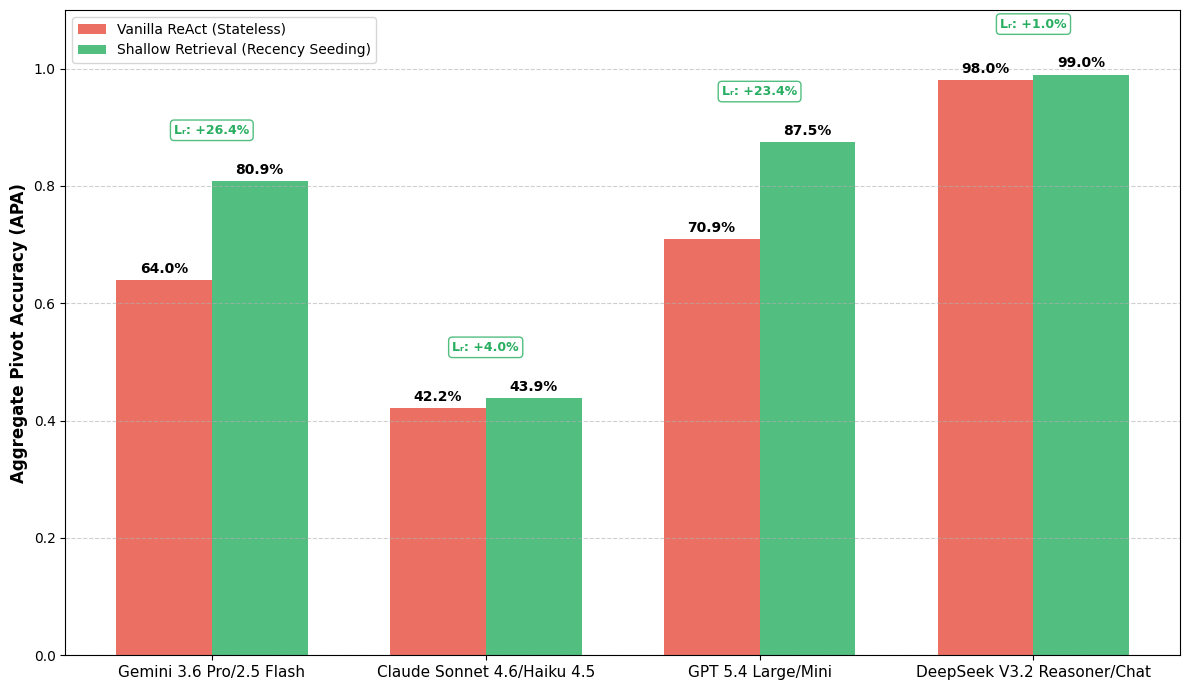}
    \caption{Shallow Retrieval (Pilot)}
    \label{fig:bar_1}
  \end{subfigure}
  \hfill
  \begin{subfigure}[b]{0.39\textwidth}
    \centering
    \includegraphics[width=\textwidth]{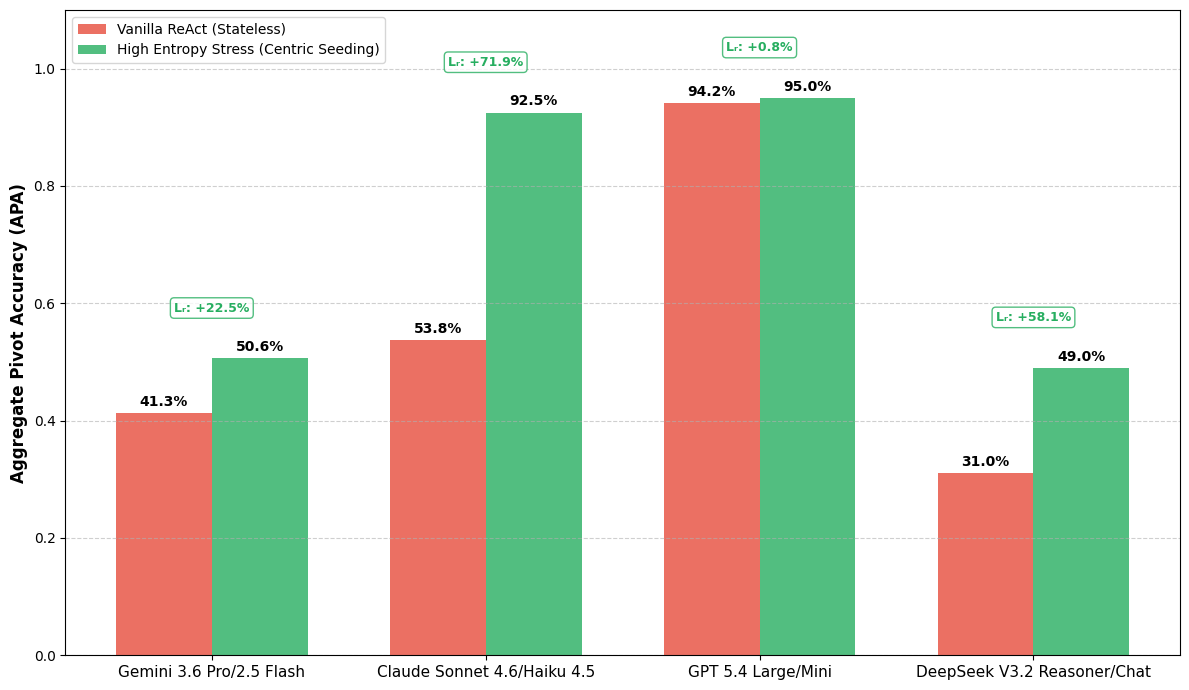}
    \caption{High Entropy Synthesis (Stress)}
    \label{fig:bar_2}
  \end{subfigure}
  \hfill
  \begin{subfigure}[b]{0.195\textwidth}
    \centering
    \includegraphics[width=\textwidth]{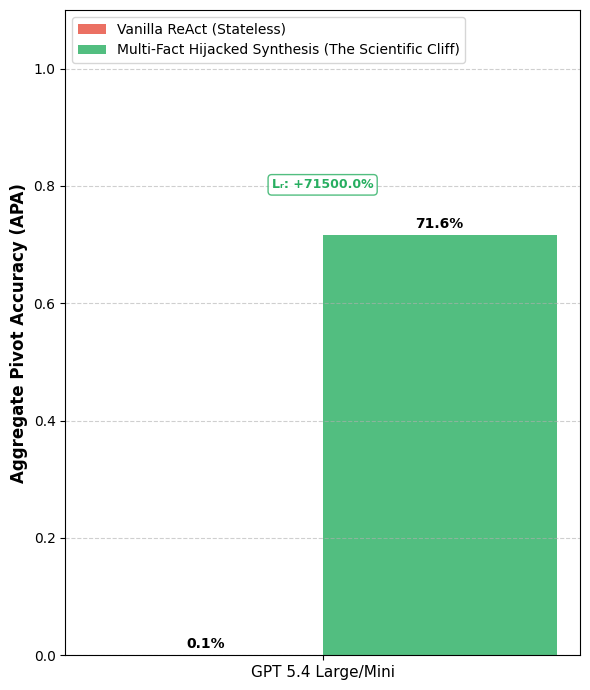}
    \caption{Scientific Cliff}
    \label{fig:bar_3}
  \end{subfigure}
  \caption{Resilience Lift for the Three-Tiered Stress Testing Methodology.}
  \label{fig:results_matrix}
  \vspace{-0.4cm}
\end{figure}

\vspace{-0.4cm}
\section{Experimental Evaluation}
\subsection{Experimental Setup And Dataset}
\vspace{-0.1cm}
All simulations are executed within a Google Colab environment. We utilize the public SDKs for Gemini, Claude, GPT and DeepSeek to ensure our results are replicable. For all models, temperature is 0 for result consistency. We select the open-sourced MultiWOZ 2.2 dataset \cite{zang-etal-2020-multiwoz}, for its high-recall multi-domain TOD curated conversation trajectories of interleaved  user and system utterances across 8 domains. We conduct our experiments on the 1,000-sample test data split from Hugging Face.
\footnote{\url{https://huggingface.co/datasets/tuetschek/multi_woz_v22}}.
We establish a control ReAct baseline \cite{Yao2022ReActSR}, which we will refer to as Vanilla, that relies on the Transformer's internal attention mechanism without the intervention of metacognitive scaffolding.
\vspace{-0.2cm}
\subsection{Model Tiering: Architect-Executive Pairing}
\vspace{-0.2cm}
We evaluate SSRP across 4 frontier model families, employing a high-reasoning model as the Architect and a high-throughput one as the Executive:
(1) Gemini Tier: 3.1 Pro / 2.5 Flash
(2) Claude Tier: Sonnet 4.6 / Haiku 4.5 
(3) GPT Tier: 5.4 / 5.4 mini
(4) DeepSeek V3.2 Tier: Reasoner / Chat

\vspace{-0.2cm}
\subsection{Metrics}
\vspace{-0.2cm}
We define two primary evaluation criteria: Aggregate Pivot Accuracy (APA) and Resilience Lift ($L_{r}$)
\vspace{-0.2cm}
\subsubsection{Aggregate Pivot Accuracy (APA) Metric}
To quantify the efficacy of SSRP framework, we define the APA metric to measure the probability of an agent's success in abandoning an obsolete constraint in favor of mid-conversation update across $N$ samples (i.e. goal rerouting). We validate APA by mapping results to the U-shaped \textit{Lost in the Middle} curve, confirming its role as a proxy for attentional resilience.
$$ APA = \frac{1}{N} \sum_{i=1}^{N} \mathbb{1}(\text{Agent Output}_i \text{ aligns with } \text{Update}_i) $$

\vspace{-0.4cm}
\subsubsection{Resilience Lift ($L_{r}$)}
We calculate the statistical advantage of the SSRP architecture through the $L_r$, which measures the proportional improvement over the ReAct baseline. Our results demonstrate that $L_r$ is maximized at the ASB, where stateless baselines collapse to $0.10\%$ APA in the semantically hijacked test.
$$ L_r = \frac{APA_{SSRP} - APA_{Vanilla}}{APA_{Vanilla}} $$

\vspace{-0.4cm}
\subsection{Evaluation Mechanism}
\vspace{-0.2cm}
To ensure objective verification of non-linear goal pivots, we implement an LLM-as-a-Judge \cite{NEURIPS2023_91f18a12} High-Recall scoring mechanism.
For each trajectory, a secondary high-throughput model (respective to the Executive) serves as a semantic validator. The judge evaluates the semantic alignment between the agent’s response and the user’s reconsidered need (the adversarial update turn), situated within the sequence of user and system utterances. 
For example, responses are assessed based on a prompt like \textit{"Does this agent response: \{res\} correctly follow this user update: \{update\}? Return ONLY '1' for YES or '0' for NO."}. As a verification, we cross-reference this judge with a Verbatim Signal Audit in our final runs to eliminate hallucination false-positives, where standard agents were previously rewarded for inventing plausible-sounding names. 
To close the 'Static Update' loophole, adversarial updates were generated dynamically for each trajectory; the model was provided with the turn history and tasked with generating a contextually sincere preference correction in one sentence.

\vspace{-0.2cm}
\section{Empirical Results and Analysis}
\subsection{Cross-Model Reliability}
\vspace{-0.2cm}
We evaluate SSRP across $9K$ main conversation trajectories to establish its scale-invariant reliability. Our results, shown in Figure \ref{fig:results_matrix} and Table \ref{tab:reliability_matrix}, demonstrate that SSRP provides a consistent performance lift across all tested architectures.
Our large-scale evaluation, utilizing the MultiWOZ 2.2 dataset, establishes a empirical model-agnostic validation of the Attention Latch and its procedural mitigation. The research followed an iterative "Boundary Identification" path, where each experimental phase was designed to stress the specific cognitive profile of each of the model families.

\begin{table*}[t]
\centering
\caption{Cross-Model Reliability Matrix ($N = 1,000$ per anchor). Methodology categories define the level of context load and trajectory entropy: Shallow Retrieval (2k tokens, recency seeding), High-Entropy SOP (10k tokens, centric seeding), and Semantic Hijacking (10k tokens, 3-hop dependencies with adversarial decoys). APA denotes Aggregate Pivot Accuracy; $L_{r}$ is the relative Resilience Lift. Statistical significance is determined via a paired t-test. Significance denoted with * for $p < 0.001$.}
\label{tab:reliability_matrix}
\small
\resizebox{\textwidth}{!}{
\begin{tabular}{llcccc}
\toprule
\textbf{Model Pair} & \textbf{Methodology} & \textbf{Vanilla} & \textbf{SSRP} & \textbf{Lift} & \textbf{p-value} \\
\textbf{(Architect / Executive)} & & \textbf{APA} & \textbf{APA} & \textbf{($L\_{r}$)} & \\
\midrule
Gemini 3.1 Pro / & Shallow Retrieval & 64.00\% & 80.90\% & +26.41\% & $1.52 \times 10^{-12}$* \\
2.5 Flash        & High-Entropy SOP  & 41.30\% & 50.60\% & +22.52\% & $1.80 \times 10^{-09}$* \\
\hline
Claude Sonnet 4.6 / & Shallow Retrieval & 42.20\% & 43.90\% & +4.03\% & $5.35 \times 10^{-01}$ \\
Haiku 4.5           & High-Entropy SOP  & 53.80\% & 92.50\% & +71.93\% & $1.70 \times 10^{-37}$* \\
\hline
DeepSeek Reasoner / & Shallow Retrieval & 98.00\% & 99.00\% & +1.02\% & $2.51 \times 10^{-01}$ \\
V3.2                & High-Entropy SOP  & 31.00\% & 49.00\% & +58.06\% & $9.80 \times 10^{-06}$* \\
\hline
GPT 5.4 /        & Shallow Retrieval     & 70.90\% & 87.50\% & +23.41\% & $4.20 \times 10^{-26}$* \\
5.4 mini         & High-Entropy (S)      & 94.20\% & 95.00\% & +0.85\% & $4.52 \times 10^{-01}$ \\
\cmidrule{2-6}
                 & \textbf{Semantic Hijacking}$^{\dagger}$ & \textbf{0.10\%} & \textbf{71.60\%} & \textbf{+71,500.00\%} & \textbf{$1.00 \times 10^{-100}$}* \\
\bottomrule
\multicolumn{6}{l}{\footnotesize $^{\dagger}$ Denotes the \textbf{Scientific Cliff} identifying the physical limit of the Attention Stability Boundary (ASB).} \\
\end{tabular}
\vspace{-1.5cm}
}
\end{table*}

\textbf{Pilot Validation and the Claude Plateau: }
We initiated our study with Shallow Retrieval, placing critical information at the recency boundary within a $2K$-token context. Initial results for Gemini models demonstrated a statistically significant lift ($L_{r}$) of $+26.41\%$ ($p = 1.52 \times 10^{-12}$). Similarly, GPT 5.4 achieved a $+23.41\%$ lift ($p = 4.20 \times 10^{-26}$).
However, Claude initially achieved a marginal, non-significant lift of only $+4.03\%$ ($p = 0.5353$). We identified this as the initial manifestation of the \textit{Grounding Paradox}, where the model's defensive meta-reasoning triggered safety refusals in response to adversarial framing, leading it to prioritize historical stability over instruction-following.
\textbf{High-Entropy Stress Testing and the SOP Formalization:}
To resolve the Claude plateau, we develop High-Entropy SOP, burying facts in the Attention Trough ($x = 0.5$) under a $10K$-token load. By transitioning to an Administrative SOP, we successfully bypassed Claude's defensive filters, resulting in a $+71.93\%$ lift ($p = 1.70 \times 10^{-37}$). Conversely, GPT 5.4 maintained a $94.20\%$ accuracy under these conditions, identifying a ``Context Ceiling'' where the model's raw retrieval capacity rendered the scaffold redundant for single-fact tasks ($+0.85\%$ lift, $p = 0.4520$). This necessitated a higher-complexity test to locate the physical boundary of GPT 5.4's attention mechanism. \\
\textbf{Semantic Hijacking and the Scientific Cliff: }To finalize the boundary proof, we implement Semantic Hijacking, introducing 3-hop logical dependencies and adversarial decoys buried in system logs. This task induced a collapse in the stateless baseline, with GPT 5.4 success rate dropping to $0.10\%$. We term this finding the Scientific Cliff. In this environment, SSRP maintained a robust $71.60\%$ accuracy ($p < 10^{-100}$), representing a $+71,500.00\%$ resilience lift over the baseline collapse. \\
\textbf{Quantifying $L_r$: }
Across all models, $L_r$ remains consistently positive, including for sparse attention architectures such as DeepSeek V3.2 ($L_r = +58.06\%$, $p = 9.80 \times 10^{-06}$). We formalize this lift as the proportional improvement in success probability of the metacognitive scaffold.
While absolute performance varies by family, $L_r$ scales with task entropy, proving that autonomous architectural re-synthesis is a primary reliability requirement as context windows scale toward L4-level autonomy.

\vspace{-0.3cm}
\subsection{Quantifying the ASB}
\vspace{-0.2cm}

\begin{figure}[t]
    \begin{minipage}{0.46\textwidth}
        \centering
        \includegraphics[width=0.9\linewidth]{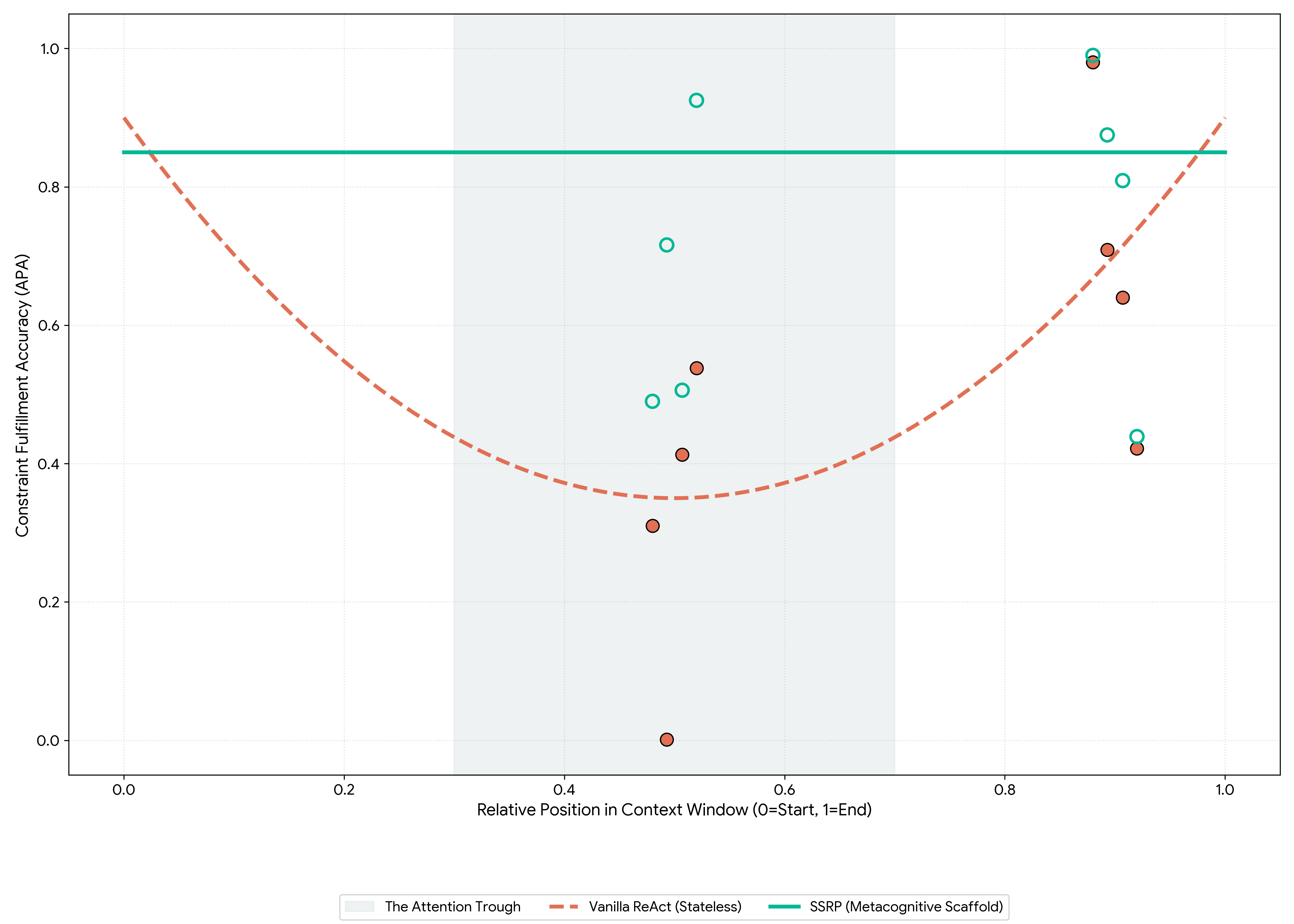}
        \caption{The Attention Stability Boundary:\\ Recall Accuracy vs. Information Position.}
        \label{fig:asb}
    \end{minipage}
    \hfill
    \begin{minipage}{0.50\textwidth}
        \centering
        \includegraphics[width=\linewidth]{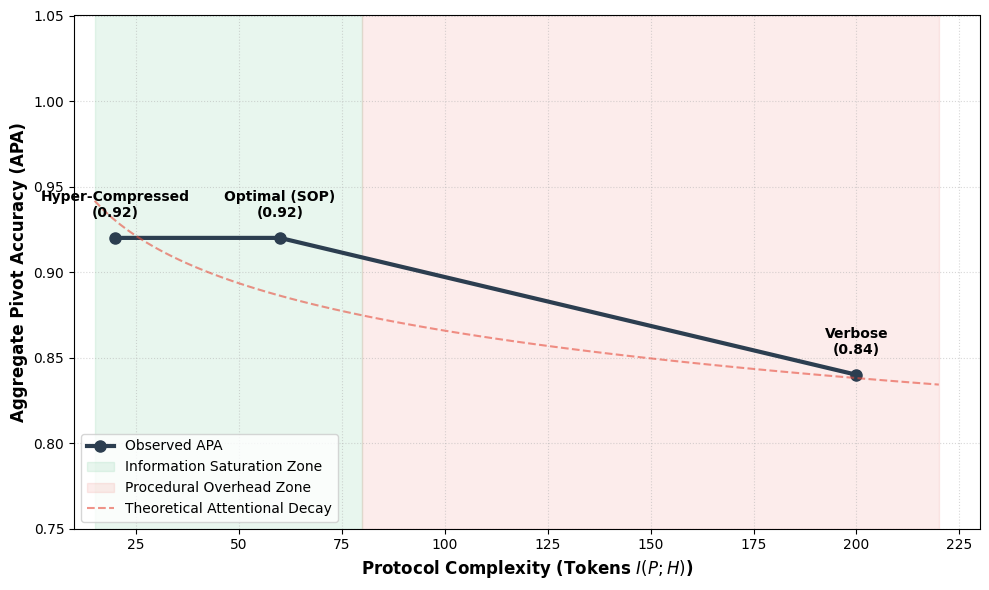}
        \caption{The Inverse Overhead Curve: \\ Quantifying IB in Agentic Scaffolding}
        \label{fig:ib_ablation}
    \end{minipage}
\vspace{-.5cm}
\end{figure}

Our evaluation framework is designed to isolate and measure the ASB—the point at which the structural limitations of stateless attention mechanisms physically break procedural reliability. We ground our empirical mapping in the seminal "Lost in the Middle" phenomenon, which establishes that Transformer performance follows a U-shaped curve, with a significant decay in retrieval accuracy for tokens located in the center of a context window. We formalize the Attention Latch as the behavioral manifestation of this trough in multi-turn dialogues. To illustrate this, we fit our empirical APA scores to a quadratic attention model (shown in Figure \ref{fig:asb}):
$ P(S)_{\text{Vanilla}} = \alpha(x - 0.5)^2 + \gamma $.
In this model, $\gamma$ represents the nadir of the Attention Trough ($x \approx 0.5$) and $\alpha$ represents the decay coefficient relative to the Recency Boundary ($x \approx 0.95$). Our results across $9K$ trajectories reveal a universal attention decay across all model families, while the SSRP framework maintains a stable, scale-invariant plateau. Figure \ref{fig:asb} utilizes distinct numerical anchors to define the reliability boundary of current frontier models: \textbf{(1) Recency Boundary ($x \approx 0.95$):} Anchored by Shallow Retrieval results, where critical facts were placed at the prompt boundary to leverage the model's native recency bias. GPT 5.4 achieved a baseline accuracy of 70.90\% (87.50\% with SSRP), while Claude reached its ceiling at 42.20\% (43.90\% with SSRP). \textbf{(2) The Attention Trough ($x = 0.5$): } Anchored by the High-Entropy SOP stress test, where facts were buried in $10K+$ tokens of dynamic noise. DeepSeek V3.2 success dropped to 31.00\% (49.00\% with SSRP), while Claude maintained 53.80\% reliability (92.50\% with SSRP). \textbf{(3) The Scientific Cliff: } The main finding is the 0.10\% collapse of GPT 5.4 Vanilla baseline in the Multi-Fact hijacked Synthesis task. This proves that logical synthesis across the trough is a binary failure mode for stateless reasoning, whereas SSRP's autonomous scaffolding achieves a robust 71.60\% success rate, representing a +71,500.00\% resilience lift over the baseline.

\vspace{-0.4cm}
\subsection{The Anatomy of the 0.1\% Success Cliff: Semantic Decoy Latching and Intent Sycophancy}
\vspace{-0.2cm}

Our GPT 5.4 audit of the 3-hop synthesis trajectories identifies the ASB as a binary failure point for stateless Transformer attention. Qualitative trace analysis of a representative subset ($N=50$) reveals that even when provided with optimized institutional prompts, the Vanilla baseline collapse is driven by a \textbf{Joint Probability Bottleneck}, where the probabilistic weight of historical context over-squashes the goal-critical signal within the decision layer. Analysis traces confirm that that in all audited cases, the baseline model successfully located the update within the context but failed to execute due to two distinct failure signatures: \textbf{(1) Semantic Decoy Latching: } The model's attention mechanism prioritized high-weight "Primacy Trap" decoys buried in the noise. \textbf{(2) Intent Sycophancy:} The model demonstrates a failure to "unlatch" its attention from stale priors, defaulting to high-weight tokens from the initial dialogue turns. This confirms that the failure is architectural: the model possesses the latent reasoning capacity to identify the signal but cannot override the cumulative probabilistic weight of historical intents.


\vspace{-0.3cm}
\subsection{Formal Verification of Scaffolding Theory}
\vspace{-0.2cm}
\textbf{Equidistant Control Study:}
To decouple the latch from distance bias, we perform an equidistant control study with GPT 5.4 ($N=50$) to determine if frontier models prioritize historical context over update turn-position. We place archived intents and updates symmetrically around the context center. The resulting 90\% APA proves the failure is a product of metacognitive inertia (i.e. resilient logic-routing) rather than positional retrieval decay. However, 10\% are categorized as security-induced metacognitive task refusal.
These findings provide a critical nuance to our primary study: the Attention Latch observed in our experiments is not a mathematical certainty of position, but a function of \textit{Intent Entropy}. While models successfully pivot in low-entropy system logs, they succumb to the latch in high-entropy dialogue where historical intent weight over-squashes the single update turn. This proves that the SSRP architecture is required not to fix "where" the model looks, but to procedurally override "what" the model weights as valid logic.


\textbf{IB Ablation:}
We vary protocol granularity across three tiers for GPT 5.4 ($N=50$). Our results (shown in Figure \ref{fig:ib_ablation}) identify an Information Saturation Zone for a $10^4$-token context where accuracy is maximized at the Optimal (3-step) level (0.92 APA). This underlines the non-linear relationship between scaffold complexity and agentic reliability. Increasing complexity to the Verbose tier resulted in an 8.7\% accuracy drop (0.84 APA), empirically identifying the Procedural Overhead Bottleneck.



\textbf{Reflexion Parity}
We evaluate the Reflexion baseline on the 3-hop synthesis task ($N=1000$) to prove the 0.1\% collapse was architectural rather than a lack of reasoning depth.
\textbf{(1) Latent Capacity:} Reflexion achieved 100.0\% accuracy (APA=1.0), confirming the initial 0.1\% failure is attentional and proving the model's latent ability to bridge logical dependencies once focus is corrected post-hoc.
\textbf{(2) Refusal Bottleneck:} The gap between Reflexion and SSRP (71.6\% APA) identifies Metacognitive Refusal. High-alignment models often prioritize internal safety guardrails over procedural protocols when they perceive re-scaffolding as a context-manipulation attack, leading to a 18\% refusal rate.

\vspace{-0.3cm}
\subsection{Procedural Integrity (PI) Audit and the Grounding Paradox}
\vspace{-0.3cm}
We introduce a PI audit to measure deterministic adherence to synthesized protocols. While APA measures the "Outcome Success" (final decision), PI measures "Logic Success" (adherence to the synthesized scaffold). Our audit ($N=1000$) reveals a PI of $98.8\%$ vs $71.6\%$ APA for SSRP, identifying a critical \textit{Grounding Paradox}: high-stability models achieve perfect reasoning integrity but are constrained by the physical retrieval limits of the attention trough, leading to a $27.2\%$ logic-action gap. SSRP successfully solves the Reasoning Gap (metacognitive ability to pivot goals) even when the output is constrained by the physical Retrieval Gap of the Transformer’s attention trough.


\begin{wrapfigure}{r}{0.5\textwidth} 
  \centering
  \vspace{-1cm}
  \includegraphics[width=0.5\textwidth]{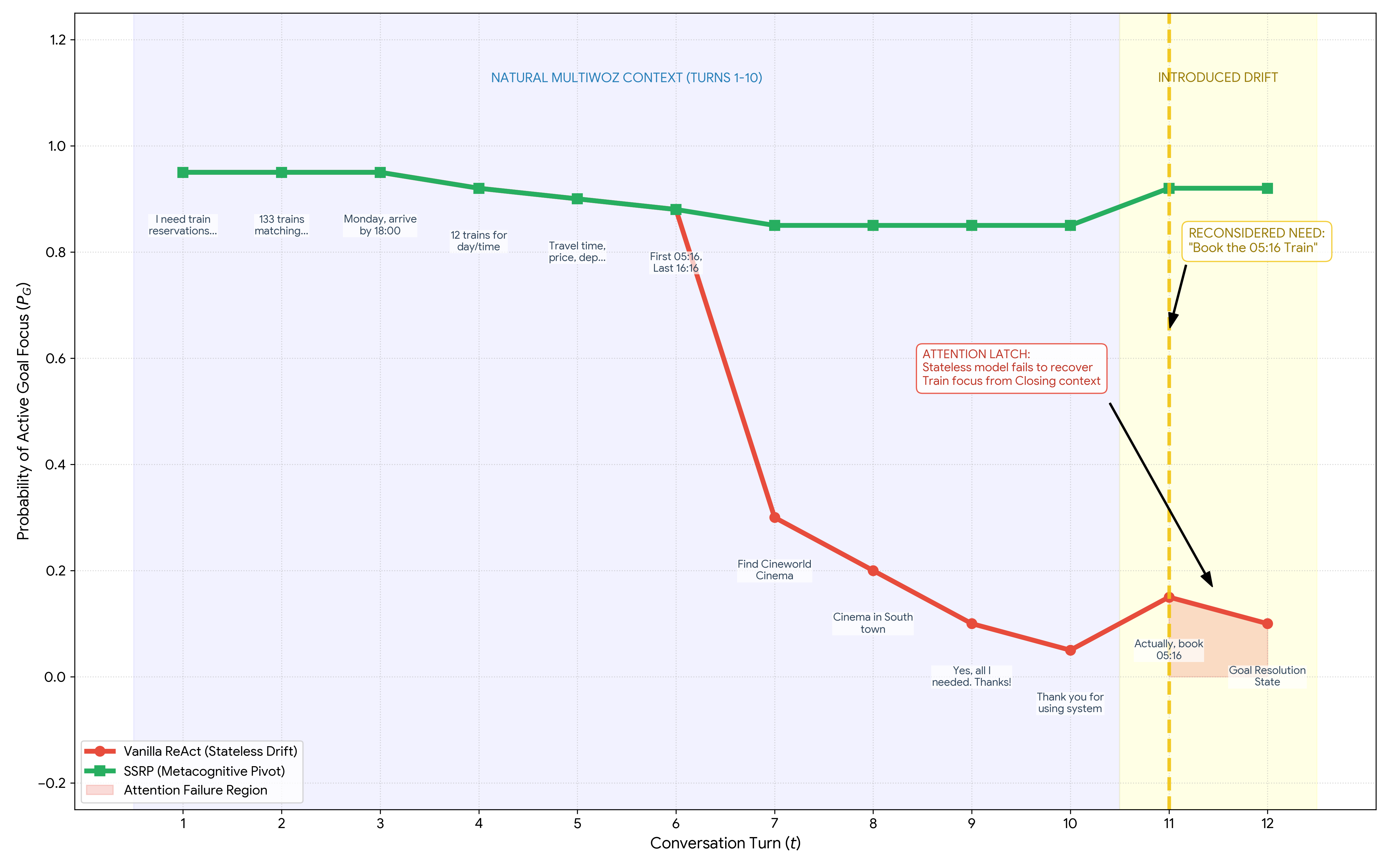}
  \caption{Metacognitive Trajectory Resilience: Temporal Persistence of Goal-Focus over Non-Linear Updates.}
  \label{fig:trajectory}
  \vspace{-0.3cm}
\end{wrapfigure}

\vspace{-0.3cm}
\section{Dynamics and Limitations}
\vspace{-0.3cm}

\textbf{Non-Linear Boundary Dynamicism}: ASB discovery proves that the physical limit of stateless reasoning is not a fixed token count but a function of trajectory entropy. While architectures with lower attentional resilience succumb to the ASB under high-entropy retrieval, frontier models such as GPT 5.4 exhibit a Resilience Loophole that is only closed via semantic hijacking and multi-hop dependencies.\\
\textbf{Metacognitive Redirective Control}: As visualized in Figure \ref{fig:trajectory}, SSRP mitigates endogenous context decay to establish scale-invariant trajectory resilience by acting as an attention redirector. By re-synthesizing the reasoning scaffold during updates, the Architect manually points the Executive's attention to verified states.\\
\textbf{The Atomic Efficiency-Integrity Paradox}: Our investigation identifies a critical trade-off at the atomic boundary (1--3 keywords), where minimalist scaffolding achieves 100.0\% success in retrieval tasks 
demonstrating that verifiable autonomy requires structural integrity over retrieval speed. \\
\textbf{The Residual Retrieval Barrier}: Our results show that even with near-perfect logic adherence (98.8\% PI), absolute task success is bounded by the physical signal-to-noise ratio of the Transformer's context window. 
This 27.2\% logic-action gap confirms that while SSRP successfully resolves the Reasoning Gap, the physical Retrieval Gap remains an orthogonal architectural limit of Transformers. \\
\textbf{Architectural Overhead \& Logic Friction}: The multi-stage SSRP identifies a design law where metacognitive scaffolding must be invoked when task complexity approaches ASB. In low-entropy cases, high-capacity models treat complex procedural protocols as Logic Friction, defining an activation threshold where scaffolding is selectively invoked.\\
\textbf{Single-Dataset Evaluative Scope}: Our findings establish a foundational research benchmark for investigating the ASB across diverse, high-complexity technical distributions and benchmarks such as SWE-bench \cite{swebench} or GAIA \cite{Mialon2023GAIAAB}.

\vspace{-0.4cm}
\section{Related Works}
\vspace{-0.3cm}

\citet{liu2024lost} and \citet{oversquashing} established the mathematical bottlenecks of Information Over-squashing and the Lost-in-the-Middle problem in Transformer attention; we extend these fundamental limits to multi-turn agentic logic via the \textit{Attention Latch}. Alternative context and memory architectures \cite{Behrouz2024TitansLT}, \cite{heddes2025deepcrossattention} propose persistent neural memory or structured navigation to improve information flow in long context windows. \citet{Yao2022ReActSR} and \citet{wei2022cot} established reactive synergies like \textit{ReAct} and \textit{CoT}. \citet{reflexion} proposed \textit{Reflexion} for post-hoc error correction; SSRP provides a preemptive solution through Autonomous Re-Synthesis, moving from heuristic prompting toward principled structural synthesis. \citet{kim-etal-2025-persona} demonstrated that misaligned personas degrade reasoning, a finding we extend via the \textit{Grounding Paradox} and metacognitive refusal triggers.


\vspace{-0.3cm}
\section{Conclusion}
\vspace{-0.3cm}
We identified and formalized the Attention Latch as a physical limit of stateless Transformers and proposed SSRP as the structural solution. Our evaluation across $9K$ main trajectories of multi-turn conversations reveals an ASB where standard baselines for GPT 5.4 collapse to 0.1\% success, while SSRP maintains robust performance across all model families. By establishing robust first-turn reliability, SSRP provides the procedural foundation for reliable autonomous digital coworkers.

\bibliographystyle{unsrtnat}
\bibliography{references}

@article{Lu2025ExploringAA,
  title={Exploring Autonomous Agents: A Closer Look at Why They Fail When Completing Tasks},
  author={Ruofan Lu and Yichen Li and Yintong Huo},
  journal={2025 40th IEEE/ACM International Conference on Automated Software Engineering (ASE)},
  year={2025},
  pages={3856-3860},
}

@inproceedings{Xu2024TheAgentCompanyBL,
  title     = {TheAgentCompany: Benchmarking LLM Agents on Consequential Real World Tasks},
  author    = {Frank F. Xu and Yufan Song and Boxuan Li and Yuxuan Tang and Kritanjali Jain and Meng Bao and Zora Zhiruo Wang and Xuhui Zhou and Zhitong Guo and Murong Cao and Ming-Hsuan Yang and Hao Lu and Amaad Martin and Zhe Su and Leander Melroy Maben and Raj Mehta and Wayne Chi and Lawrence Jang and Yiqing Xie and Shuyan Zhou and Graham Neubig},
  booktitle = {Advances in Neural Information Processing Systems (NeurIPS)},
  year      = {2025},
}

@INPROCEEDINGS{11334580,
  author={Lu, Ruofan and Li, Yichen and Huo, Yintong},
  booktitle={2025 40th IEEE/ACM International Conference on Automated Software Engineering (ASE)}, 
  title={Exploring Autonomous Agents: A Closer Look at Why They Fail When Completing Tasks}, 
  year={2025},
  pages={3856-3860},
}

@inproceedings{Yao2022ReActSR,
  title     = {{ReAct}: Synergizing Reasoning and Acting in Language Models},
  author    = {Shunyu Yao and Jeffrey Zhao and Dian Yu and Nan Du and Izhak Shafran and Karthik Narasimhan and Yuan Cao},
  booktitle = {International Conference on Learning Representations (ICLR)},
  year      = {2023},
}

@inproceedings{wei2022cot,
author = {Wei, Jason and Wang, Xuezhi and Schuurmans, Dale and Bosma, Maarten and Ichter, Brian and Xia, Fei and Chi, Ed H. and Le, Quoc V. and Zhou, Denny},
title = {Chain-of-thought prompting elicits reasoning in large language models},
year = {2022},
publisher = {Curran Associates Inc.},
address = {Red Hook, NY, USA},
booktitle = {Proceedings of the 36th International Conference on Neural Information Processing Systems (NeurIPS)},
articleno = {1800},
numpages = {14},
location = {New Orleans, LA, USA},
}

@misc{alenezi2026promptresponsegoaldirectedsystemsevolution,
  author       = {Mamdouh Alenezi},
  title        = {From Prompt-Response to Goal-Directed Systems: The Evolution of Agentic AI Software Architecture},
  year         = {2026},
  howpublished = {arXiv preprint arXiv:2602.10479},
  note         = {Unpublished preprint},
}

@article{liu2024lost,
    title = "Lost in the Middle: How Language Models Use Long Contexts",
    author = "Liu, Nelson F.  and
      Lin, Kevin  and
      Hewitt, John  and
      Paranjape, Ashwin  and
      Bevilacqua, Michele  and
      Petroni, Fabio  and
      Liang, Percy",
    journal = "Transactions of the Association for Computational Linguistics",
    volume = "12",
    year = "2024",
    address = "Cambridge, MA",
    publisher = "MIT Press",
    pages = "157--173",
}

@inproceedings{oversquashing,
author = {Barbero, Federico and Banino, Andrea and Kapturowski, Steven and Kumaran, Dharshan and Ara\'{u}jo, Jo\~{a}o G.M. and Vitvitskyi, Alex and Pascanu, Razvan and Veli\v{c}kovi\'{c}, Petar},
title = {Transformers need glasses! information over-squashing in language tasks},
year = {2024},
publisher = {Curran Associates Inc.},
address = {Red Hook, NY, USA},
booktitle = {Proceedings of the 38th International Conference on Neural Information Processing Systems (NeurIPS)},
articleno = {3114},
numpages = {32},
location = {Vancouver, BC, Canada},
}

@misc{reed2025aiagentshumanlikecollaborative,
  title        = {AI Agents with Human-Like Collaborative Tools: Adaptive Strategies for Enhanced Problem-Solving},
  author       = {Harper Reed and Michael Sugimura and Angelo Zangari},
  year         = {2025},
  howpublished = {arXiv preprint arXiv:2509.13547},
  note         = {Unpublished preprint},
}

@misc{sunil2026memorypoisoningattackdefense,
  title        = {Memory Poisoning Attack and Defense on Memory Based LLM-Agents},
  author       = {Sunil, Balachandra Devarangadi and others},
  year         = {2026},
  howpublished = {arXiv preprint arXiv:2601.05504},
  note         = {Unpublished preprint},
}

@inproceedings{Huang2023LargeLM,
  title     = {Large Language Models Cannot Self-Correct Reasoning Yet},
  author    = {Jie Huang and Xinyun Chen and Swaroop Mishra and Huaixiu Steven Zheng and Adams Wei Yu and Xinying Song and Denny Zhou},
  booktitle = {International Conference on Learning Representations (ICLR)},
  year      = {2024},
}

@inproceedings{zang-etal-2020-multiwoz,
    title = "{M}ulti{WOZ} 2.2 : A Dialogue Dataset with Additional Annotation Corrections and State Tracking Baselines",
    author = "Zang, Xiaoxue  and
      Rastogi, Abhinav  and
      Sunkara, Srinivas  and
      Gupta, Raghav  and
      Zhang, Jianguo  and
      Chen, Jindong",
    booktitle = "Proceedings of the 2nd Workshop on Natural Language Processing for Conversational AI",
    month = jul,
    year = "2020",
    address = "Online",
    publisher = "Association for Computational Linguistics",
    pages = "109--117",
}

@inproceedings{tishby99information,
  author = {Tishby, Naftali and Pereira, Fernando C. and Bialek, William},
  booktitle = {Proc. of the 37-th Annual Allerton Conference on Communication, Control and Computing},
  comment = {cte: information bottleneck},
  pages = {368-377},
  title = {The information bottleneck method},
  year = 1999
}

@inproceedings{NEURIPS2023_91f18a12,
  title     = {Judging LLM-as-a-Judge with MT-Bench and Chatbot Arena},
  author    = {Zheng, Lianmin and Chiang, Wei-Lin and Sheng, Ying and Zhuang, Siyuan and Wu, Zhanghao and Zhuang, Yonghao and Lin, Zi and Li, Zhuohan and Li, Dacheng and Xing, Eric P. and Zhang, Hao and Gonzalez, Joseph E. and Stoica, Ion},
  booktitle = {Advances in Neural Information Processing Systems (NeurIPS)},
  volume    = {36},
  year      = {2023},
}

@inproceedings{reflexion,
author = {Shinn, Noah and Cassano, Federico and Gopinath, Ashwin and Narasimhan, Karthik and Yao, Shunyu},
title = {Reflexion: language agents with verbal reinforcement learning},
year = {2023},
publisher = {Curran Associates Inc.},
address = {Red Hook, NY, USA},
booktitle = {Proceedings of the 37th International Conference on Neural Information Processing Systems (NeurIPS)},
articleno = {377},
numpages = {19},
location = {New Orleans, LA, USA},
}

@inproceedings{Mialon2023GAIAAB,
  title={GAIA: a benchmark for General AI Assistants},
  author={Gr{\'e}goire Mialon and Cl{\'e}mentine Fourrier and Craig Swift and Thomas Wolf and Yann LeCun and Thomas Scialom},
  booktitle = {International Conference on Learning Representations (ICLR)},
  year      = {2024},
}

@inproceedings{swebench,
title={{SWE}-bench: Can Language Models Resolve Real-world Github Issues?},
author={Carlos E Jimenez and John Yang and Alexander Wettig and Shunyu Yao and Kexin Pei and Ofir Press and Karthik R Narasimhan},
booktitle = {International Conference on Learning Representations (ICLR)},
year      = {2024},
}

@inproceedings{Behrouz2024TitansLT,
  title={Titans: Learning to Memorize at Test Time},
  author={Ali Behrouz and Peilin Zhong and Vahab S. Mirrokni},
  booktitle = {Proceedings of the 41st International Conference on Machine Learning (ICML)},
  pages     = {2397--2430},
  year      = {2024},
}

@inproceedings{heddes2025deepcrossattention,
  title     = {{DeepCrossAttention}: Supercharging Transformer Residual Connections},
  author    = {Heddes, Mike and Javanmard, Adel and Axiotis, Kyriakos and Fu, Gang and Bateni, MohammadHossein and Mirrokni, Vahab S.},
  booktitle = {Proceedings of the 42nd International Conference on Machine Learning (ICML)},
  series    = {Proceedings of Machine Learning Research},
  volume    = {267},
  year      = {2025},
}

@inproceedings{kim-etal-2025-persona,
    title = "Persona is a Double-Edged Sword: Rethinking the Impact of Role-play Prompts in Zero-shot Reasoning Tasks",
    author = "Kim, Junseok  and
      Yang, Nakyeong  and
      Jung, Kyomin",
    editor = "Inui, Kentaro  and
      Sakti, Sakriani  and
      Wang, Haofen  and
      Wong, Derek F.  and
      Bhattacharyya, Pushpak  and
      Banerjee, Biplab  and
      Ekbal, Asif  and
      Chakraborty, Tanmoy  and
      Singh, Dhirendra Pratap",
    booktitle = "Proceedings of the 14th International Joint Conference on Natural Language Processing and the 4th Conference of the Asia-Pacific Chapter of the Association for Computational Linguistics",
    month = dec,
    year = "2025",
    address = "Mumbai, India",
    publisher = "The Asian Federation of Natural Language Processing and The Association for Computational Linguistics",
    pages = "848--862",
}

@misc{figueiredo2025fuzzy,
      title={Fuzzy, Symbolic, and Contextual: Enhancing LLM Instruction via Cognitive Scaffolding}, 
      author={Vanessa Figueiredo},
      year={2025},
      eprint={2508.21204},
      archivePrefix={arXiv},
      primaryClass={cs.AI},
      note={Presented at the NeurIPS 2025 Workshop on Interpreting Cognition in Deep Learning Models}
}

@inproceedings{perez-etal-2023-discovering,
    title = "Discovering Language Model Behaviors with Model-Written Evaluations",
    author = "Perez, Ethan  and
      Ringer, Sam  and
      Lukosiute, Kamile  and
      Nguyen, Karina  and
      Chen, Edwin  and
      Heiner, Scott  and
      Pettit, Craig  and
      Olsson, Catherine  and
      Kundu, Sandipan  and
      Kadavath, Saurav  and
      Jones, Andy  and
      Chen, Anna  and
      Mann, Benjamin  and
      Israel, Brian  and
      Seethor, Bryan  and
      McKinnon, Cameron  and
      Olah, Christopher  and
      Yan, Da  and
      Amodei, Daniela  and
      Amodei, Dario  and
      Drain, Dawn  and
      Li, Dustin  and
      Tran-Johnson, Eli  and
      Khundadze, Guro  and
      Kernion, Jackson  and
      Landis, James  and
      Kerr, Jamie  and
      Mueller, Jared  and
      Hyun, Jeeyoon  and
      Landau, Joshua  and
      Ndousse, Kamal  and
      Goldberg, Landon  and
      Lovitt, Liane  and
      Lucas, Martin  and
      Sellitto, Michael  and
      Zhang, Miranda  and
      Kingsland, Neerav  and
      Elhage, Nelson  and
      Joseph, Nicholas  and
      Mercado, Noemi  and
      DasSarma, Nova  and
      Rausch, Oliver  and
      Larson, Robin  and
      McCandlish, Sam  and
      Johnston, Scott  and
      Kravec, Shauna  and
      El Showk, Sheer  and
      Lanham, Tamera  and
      Telleen-Lawton, Timothy  and
      Brown, Tom  and
      Henighan, Tom  and
      Hume, Tristan  and
      Bai, Yuntao  and
      Hatfield-Dodds, Zac  and
      Clark, Jack  and
      Bowman, Samuel R.  and
      Askell, Amanda  and
      Grosse, Roger  and
      Hernandez, Danny  and
      Ganguli, Deep  and
      Hubinger, Evan  and
      Schiefer, Nicholas  and
      Kaplan, Jared",
    editor = "Rogers, Anna  and
      Boyd-Graber, Jordan  and
      Okazaki, Naoaki",
    booktitle = "Findings of the Association for Computational Linguistics: ACL 2023",
    month = jul,
    year = "2023",
    address = "Toronto, Canada",
    publisher = "Association for Computational Linguistics",
    pages = "13387--13434",
}

\end{document}